\documentclass[letterpaper]{article} 
\usepackage{aaai24}  

\usepackage{times}  
\usepackage{helvet}  
\usepackage{courier}  
\usepackage[hyphens]{url}  
\usepackage{graphicx} 
\usepackage{natbib}  
\usepackage{caption} 
\usepackage{algorithm}
\usepackage{algorithmic}

\usepackage{newfloat}
\usepackage{listings}
\DeclareCaptionStyle{ruled}{labelfont=normalfont,labelsep=colon,strut=off} 
\usepackage{adjustbox}
\usepackage{multirow}
\usepackage{amsfonts}
\usepackage{amsmath}
\usepackage{booktabs}

\usepackage{footmisc}

\floatstyle{ruled}
\newfloat{listing}{tb}{lst}{}
\floatname{listing}{Listing}
\title{Fully-Connected Spatial-Temporal Graph for Multivariate Time-Series Data}
\author{
    Yucheng Wang\textsuperscript{\rm 1,3}, Yuecong Xu\textsuperscript{\rm 1}, Jianfei Yang\textsuperscript{\rm 3}, \\Min Wu\textsuperscript{\rm 1}, Xiaoli Li\textsuperscript{\rm 1,2,3}, Lihua Xie\textsuperscript{\rm 3}, Zhenghua Chen\textsuperscript{\rm 1,2}\footnote{Corresponding Author}\\
}
\affiliations{
 \textsuperscript{\rm 1}Institute for Infocomm Research, A*STAR, Singapore\\
     \textsuperscript{\rm 2}Centre for Frontier AI Research, A*STAR, Singapore\\
    \textsuperscript{\rm 3}Nanyang Technological University, Singapore\\
    \{yucheng003, xuyu0014, yang0478, chen0832\}@e.ntu.edu.sg,
    \{wumin, xlli\}@i2r.a-star.edu.sg,
    elhxie@ntu.edu.sg
}

\begin{document}

\maketitle

\begin{abstract}

Multivariate Time-Series (MTS) data is crucial in various application fields. With its sequential and multi-source (multiple sensors) properties, MTS data inherently exhibits Spatial-Temporal (ST) dependencies, involving temporal correlations between timestamps and spatial correlations between sensors in each timestamp. To effectively leverage this information, Graph Neural Network-based methods (GNNs) have been widely adopted. However, existing approaches separately capture spatial dependency and temporal dependency and fail to capture the correlations between Different sEnsors at Different Timestamps (DEDT). Overlooking such correlations hinders the comprehensive modelling of ST dependencies within MTS data, thus restricting existing GNNs from learning effective representations. To address this limitation, we propose a novel method called Fully-Connected Spatial-Temporal Graph Neural Network (FC-STGNN), including two key components namely FC graph construction and FC graph convolution. For graph construction, we design a decay graph to connect sensors across all timestamps based on their temporal distances, enabling us to fully model the ST dependencies by considering the correlations between DEDT. Further, we devise FC graph convolution with a moving-pooling GNN layer to effectively capture the ST dependencies for learning effective representations. Extensive experiments show the effectiveness of FC-STGNN on multiple MTS datasets compared to SOTA methods. The code is available at https://github.com/Frank-Wang-oss/FCSTGNN.


\end{abstract}

\section{Introduction}

Multivariate Time-Series (MTS) data has gained popularity owing to their extensive utilization in various real-world applications such as predictive maintenance and healthcare \cite{gupta2020approaches,yang2022deep}. Considering its sequential property together with multiple data sources, e.g., sensors, MTS data exhibits Spatial-Temporal (ST) dependencies, including temporal correlations between timestamps and spatial correlations between sensors in each timestamp. Traditional approaches mainly focus on capturing temporal dependencies by employing temporal encoders, disregarding spatial dependencies and thus limiting their ability to learn effective representations \cite{tan2020data,9300195}. To address this limitation, Graph Neural Network-based methods (GNNs) have emerged as popular solutions to exploit ST dependencies within MTS data \cite{wang2023sensor,ijcai2020-184}.

\begin{figure}[htbp!]
    \centering\includegraphics[width = .92\linewidth]{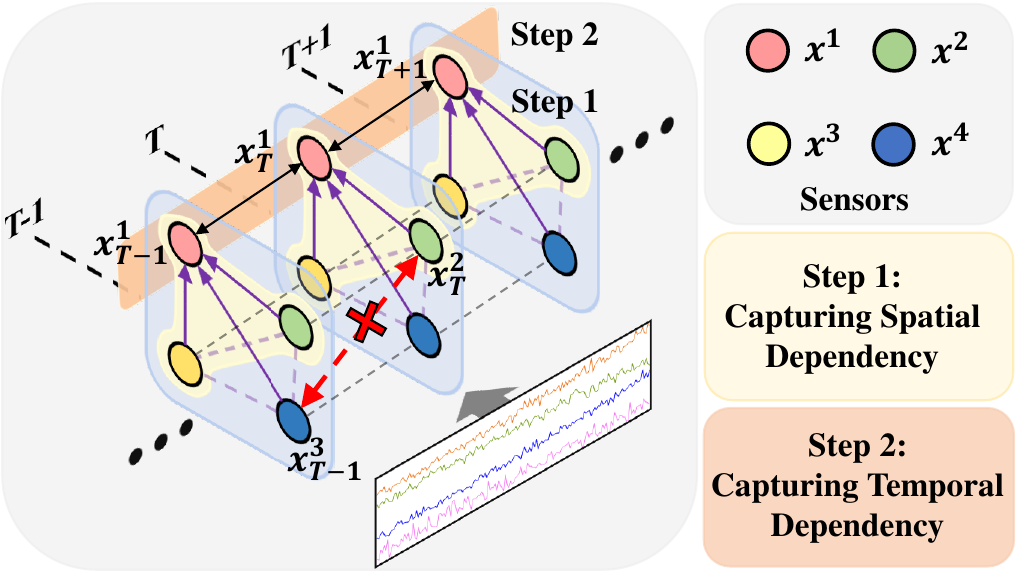}
    \caption{ST graphs are constructed from MTS data, creating separate graphs for each timestamp, to capture ST dependencies. In step 1, GNN captures the spatial dependency within each graph, e.g., [$x^1_{T-1}$, $x^2_{T-1}$, $x^3_{T-1}$]. In step 2, temporal encoders capture temporal dependencies for the corresponding sensors across different timestamps, e.g., [$x^2_{T-1}$, $x^2_{T}$, $x^2_{T+1}$]. However, this method overlooks the correlations between different sensors at different timestamps, e.g., $x^3_{T-1}$ and $x^2_T$, failing to model comprehensive ST dependencies.}
    \label{fig:nece}
\end{figure}
GNNs are always combined with temporal encoders to capture ST dependencies. The process begins with the construction of ST graphs, where separate graphs are constructed for each timestamp, representing the relationships between sensors over both time and space. 
To capture the ST dependencies, existing methods \cite{deng2021graph, wang2023sensor} primarily adopt a two-step approach, incorporating GNNs and temporal encoders to capture the spatial dependency and temporal dependency separately. As shown in Fig. \ref{fig:nece}, GNNs are initially employed to capture spatial dependencies between sensors at each timestamp, and then temporal encoders capture temporal dependencies for corresponding sensors across different timestamps (The order might be reversed).

These works have shown improved performance compared to conventional methods using temporal encoders alone. However, they process each graph independently, overlooking the correlations between Different sEnsors at Different Timestamps (DEDT), e.g., the correlation between $x^2_T$ and $x^3_{T-1}$ in Fig. \ref{fig:nece}. These correlations are crucial in modelling comprehensive ST dependencies within MTS data. For instance, we consider a machine health detection scenario where a temperature sensor is highly correlated with a fan speed sensor. In this case, not only are the two sensors at the same timestamp highly correlated, but the past temperature would also influence the future fan speed, resulting in the correlations between DEDT. Due to limitations in graph construction and graph convolution, existing methods fail to effectively capture the correlations between DEDT, restricting their ability to model the comprehensive ST dependencies within MTS data.

To solve the above limitation, we propose a novel method called Fully-Connected Spatial-Temporal Graph Neural Network (FC-STGNN), which consists of two key components: FC graph construction and FC graph convolution, together to capture the comprehensive ST dependencies within MTS data. For graph construction, we introduce an FC graph to establish full connections between all sensors across all timestamps, enabling us to fully model the ST dependencies within MTS data by additionally considering the correlations between DEDT.
The process begins by segmenting each MTS sample into multiple patches, each corresponding to a timestamp, and then encoding the signals of each sensor as sensor features. The sensors across all patches are fully connected through dot-product computations. To improve the FC graph, we design a decay matrix by considering the temporal distances between these patches, assigning larger correlations to closer patches. This design ensures that the temporally close sensors exhibit stronger correlations compared to those that are temporally distant.

We then design FC graph convolution to effectively capture the ST dependencies within the FC graph. While a naive approach would directly perform graph convolution across the entire graph by considering all sensors across all patches, we recognize that this may fail to capture local temporal patterns within MTS data, similar to how Convolutional Neural Networks (CNNs) adopt local convolution to capture local patterns within images instead of directly stacking all pixels. Additionally, using all sensors across all patches introduces unnecessary computational costs.
To address this, we propose a moving-pooling GNN layer, which adopts moving windows with a specific size to slide along patches. Within each window, graph convolution is performed to update node features through edge propagation. Subsequently, a temporal pooling operation is applied to obtain high-level sensor features. After multiple parallel layers of moving-pooling GNN, we acquire the updated sensor features, which are then stacked and mapped to obtain final representations.

In summary, our contributions are three folds. First, we propose a fully-connected ST graph to explicitly model the correlations between sensors across all timestamps. By designing a temporal distance-based decay matrix, we improve the constructed graph, effectively modelling the comprehensive ST dependencies within MTS data. Second, we propose a moving-pooling GNN layer to effectively capture the ST dependencies from the constructed graph for learning effective representations. It introduces a moving window to consider local ST dependencies, followed by a temporal pooling operation to extract high-level features. Third, we conduct extensive experiments to show the effectiveness of our method for effectively modelling and capturing the complex ST dependencies within MTS data.

\section{Related Work}
\paragraph{Conventional methods for MTS data}
Due to the inherent sequential nature of MTS data, traditional methods primarily focused on capturing temporal correlations between timestamps. This is often achieved through leveraging temporal encoders such as CNNs, Long Short-Term Memory (LSTM), and Transformers. Initially, due to the popularity in computer vision, 1D-CNN was first applied \cite{franceschi2019unsupervised,8437249,WANG2019107,wang2023multivariate,zhang2020tapnet,eldele2021time}. These models employed 1D-CNN as encoders to extract temporal features, which were then employed for downstream tasks. Additionally, some investigations explored 2D-CNN models, treating MTS data as two-dimensional images \cite{8700425}. LSTM-based model is another branch to capture the temporal dependency from MTS data due to its ability to capture long-term dependency \cite{DU2020269,LIU2020113082,9484796}. More recently, due to its powerful attention mechanism, Transformers \cite{vaswani2017attention} become popular, and extensive transformer-based works are developed to maximize its potential to capture temporal correlations within MTS data \cite{zerveas2021transformer,wu2021autoformer,zhou2021informer}.

While these methodologies have greatly advanced MTS analysis, they overlooked the spatial dependency within MTS data which originates from its multi-source nature, i.e., signals are collected from multiple sensors. The dependency represents the spatial correlations between these sensors, which play important roles in fully modelling MTS data.
For instance, in a scenario involving machine status detection, a temperature sensor's readings would correlate with those of a fan speed sensor. Overlooking the spatial dependency restricts the ability to fully model MTS data, resulting in limited performance when learning effective representations.

\renewcommand{\dblfloatpagefraction}{.9}
\begin{figure*}[htbp!]
    \centering
    \includegraphics[width = .78\linewidth]{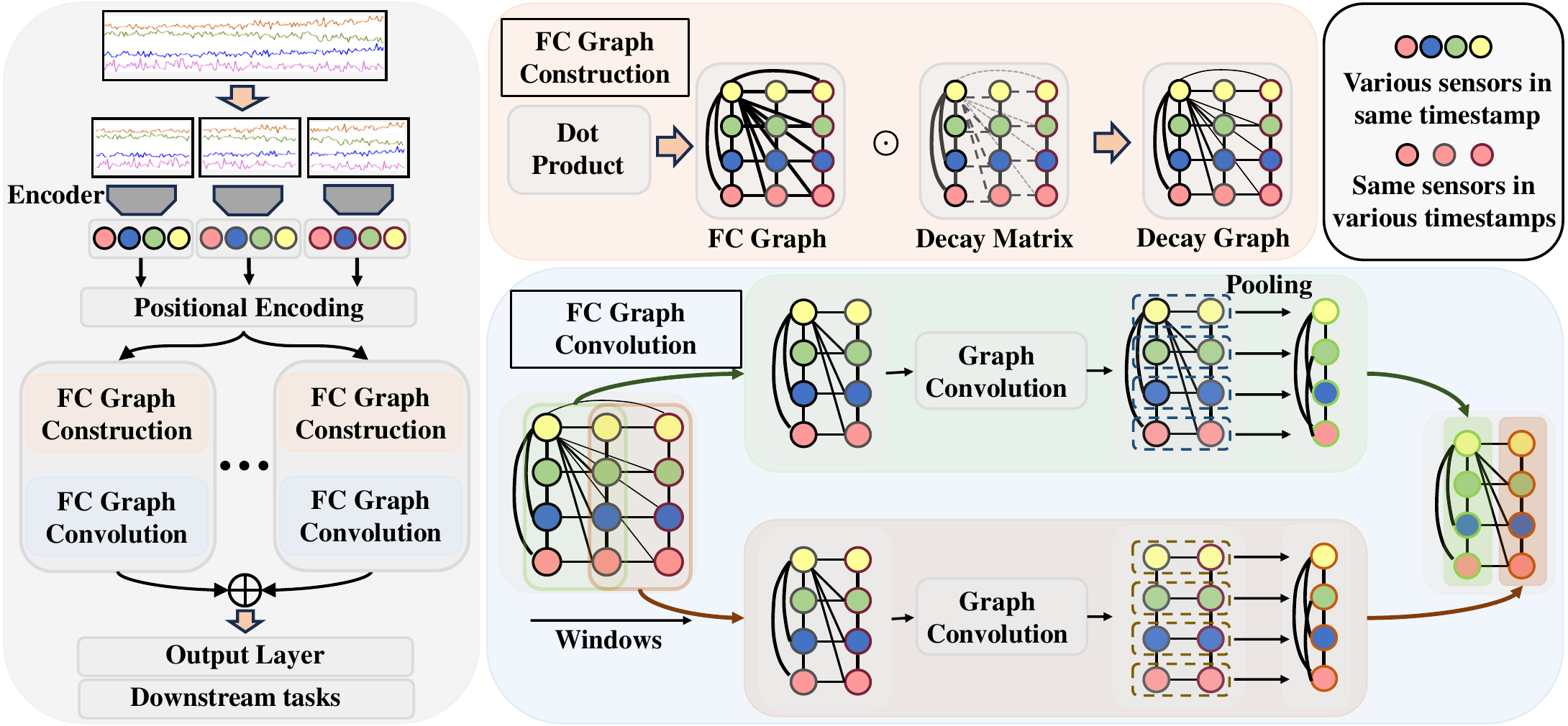}
    \caption{Overall structure of FC-STGNN. 
    Beginning with an MTS sample, each sensor's signals are segmented into multiple patches, as shown in the example with three patches (each containing four sensors). Sensor-level features are then learned through an encoder within each patch. Then, the features from different patches are further encoded with positional encoding, followed by FC graph construction and convolution.
    (1) FC graph construction: This involves fully connecting the sensors across patches by calculating their dot products, enabling the additional connections of DEDT. To refine the full connections of sensors across patches, a decay matrix is introduced by considering their temporal distances. (Note: Due to space constraints, only one sensor exhibits fully-connected weights in this example).
    (2) FC graph convolution: Moving windows with specific sizes traverse along patches (e.g., two in this example). Graph convolution is then applied to the FC graph within each window. Following the update of each sensor's features by capturing the comprehensive ST dependencies within each window, a temporal pooling operation is employed to learn high-level sensor features for each window. After multiple parallel layers, we concatenate the features, followed by an output layer to obtain final representations for downstream tasks.
}
    \label{fig:overall}
\end{figure*}

\paragraph{GNN for MTS data}

In recent years, a growing number of researchers have recognized the significance of incorporating spatial dependencies into the learning of MTS data representations \cite{jin2023survey}. To achieve that, a common approach is to leverage GNN, generally involving the combination of GNN with other temporal encoders, such as 1D-CNN, to capture the spatial dependency and temporal dependency respectively \cite{ijcai2020-184,9280401,deng2021graph,wang2023multivariate,wu2020connecting,shao2022pre,yu2017spatio}. 
For example, HierCorrPool \cite{wang2023multivariate} designed sequential graphs and adopted CNN to capture temporal dependency within these graphs. Subsequently, GNNs were utilized to capture the spatial dependencies between sensors within each graph. GraphSleepNet \cite{ijcai2020-184} also introduced sequential graphs and designed a CNN-GNN encoder to capture ST dependencies within MTS data for sleep stage classification. HAGCN \cite{LI2021107878} employed LSTM to extract temporal features, which were then used to construct graphs that were further processed by GNN. 
These researchers have made significant contributions by leveraging GNN to capture spatial dependencies within MTS data. However, as previously discussed, their approaches suffer from limitations in graph construction and graph convolution, preventing them from explicitly considering the correlations between DEDT. This limitation hinders their ability to comprehensively model ST dependencies within MTS data, ultimately impacting their performance in learning effective representations.
Similar challenges have been addressed in domains with available graph data, such as traffic and human skeleton graphs \cite{tan2023learning,song2020spatial}. However, these works typically deal with graph data containing attributed nodes. In contrast, our focus is on MTS data, where each sensor corresponds to an unattributed node, featuring only time-series signals. This distinction poses challenges when attempting to directly apply existing works designed for attributed graphs.

To address the limitation in existing approaches and comprehensively model ST dependencies within MTS data, we introduce FC-STGNN, a novel framework designed to enhance representation learning for MTS data.

\section{Methodology}

\subsection{Problem Formulation}

Given a dataset $\mathcal{D}$ consisting of $n$ labelled MTS samples $\{X_j, y_j\}_{j=1}^n$, each sample $X_j\in\mathbb{R}^{N\times{L}}$ is collected from $N$ sensors with $T$ timestamps. Our objective is to learn an effective encoder $\mathcal{F}$ capable of fully capturing the underlying spatial-temporal dependencies within MTS data. This approach can help extract effective representations $h_j = \mathcal{F}(X_j)\in\mathbb{R}^{d}$ from $X_j$, enabling us to 
perform well in diverse downstream tasks, such as machine remaining useful life prediction, human activity recognition, and so on. For simplicity, the subscript $j$ is removed, and we denote an MTS sample as $X$.

\subsection{Overall Structure}

Fig.~\ref{fig:overall} shows the overall structure of FC-STGNN, which aims to fully capture the ST dependencies within MTS data. Given an MTS sample, we first segment the signals of each sensor into multiple patches, each corresponding to a timestamp. Each patch is then processed by an encoder to learn sensor-level features. Subsequently, we employ positional encoding to integrate positional information into the sensor features across different patches. Next, we propose FC graph construction to achieve comprehensive interconnections between sensors across patches, realized by calculating the dot product of sensors. To enhance these connections, we introduce a decay matrix by considering temporal distances between patches. Next, a moving-pooling GNN is then proposed to fully capture the ST dependencies within the FC graph. We design moving windows which traverse along patches and then apply GNN within each window. After updating sensor features by capturing the comprehensive ST dependencies within each window, a temporal pooling operation is employed to learn high-level sensor features. By using multiple parallel layers of FC graph construction and convolution to capture ST dependencies from different perspectives, we concatenate the features, followed by an output layer to obtain the final representations for downstream tasks. Further details are provided in subsequent sections.

\subsection{FC Graph Construction}
\subsubsection{Graph Construction}
Given an MTS sample $X\in\mathbb{R}^{N\times{L}}$, we segment the signals of each sensor into multiple patches by considering the local temporal patterns within MTS data \cite{wang2023multivariate}. Using patch size $f$, we create $\{X_t\}_{t=1}^{\hat{L}}$ from $X$, where $t$ is the patch index representing a timestamp, and each $X_t\in\mathbb{R}^{N\times{f}}$. $\hat{L}$ denotes the number of segmented patches, calculated as $\hat{L} = [\frac{L}{f}]$, where $[\cdot]$ represents the truncation operation. Each $X_t$ contains segmented signals from $n$ sensors, i.e., $X_t = \{x_{t,i}\}_{i=1}^N$, where $x_{t,i}\in\mathbb{R}^f$.

Subsequently, we employ an encoder $f_c(\cdot|W_c)$ to process the segmented signals within each window. Notably, the encoder operates at the sensor-level to learn sensor-level features, i.e., $x'_{t,i} = f_c(x_{t,i}|W_c)$. Moreover, to maintain the directionality across patches, i.e., the relative positional information of patches, we adopt positional encoding as inspired by \cite{vaswani2017attention}. Specifically, for the $i$-th sensor $\{x'_{t,i}\}_{t=1}^{\hat{L}}$, positional encoding, as shown in Eq. (\ref{eq:posi}), is introduced into sensor features, e.g., $z_{t,i} = f_p(t) + x'_{t,i}$ representing the sensor features enhanced by positional encoding. Here, $m$ represents the $m$-th feature of sensor features.
\begin{equation}
\label{eq:posi}
    \Vec{p_t}^{(m)} = f_p(t)^{(m)} :=
    \begin{cases}
        sin(\omega_k\cdot{t}) & \text{if $m=2k$},\\
        cos(\omega_k\cdot{t}) & \text{if $m=2k+1$}.
    \end{cases}
\end{equation}

With the learned sensor features across multiple patches, we can then proceed to construct an FC graph that interconnects all sensors across these patches by additionally considering the correlations between DEDT. For graph construction, we have the assumption that correlated sensors should exhibit similar properties, making their features close within the feature space. This enables us to adopt similarity to represent the correlation between sensors, with greater similarity reflecting a higher correlation. In this case, we employ a simple yet effective metric, the dot product, to quantify the similarity between two sensors, defined as $e_{tr, ij} = g_s({z}_{t, i})(g_s({z}_{r, j}))^T$, where $t,r\in[1,\hat{L}]$ and $i,j\in[1,N]$. Here, the function $g_s(z) = zW_s$ is employed to enhance the expressive capacity, drawing inspiration from the attention computation in \cite{vaswani2017attention}, where $W_s$ is the learnable weights. Further, the softmax function restricts the correlations within [0,1]. Finally, we derive the FC graph $\mathcal{G} = (Z, E)$, where $Z = \{\{z_{t,i}\}_{i=1}^{N}\}_{t=1}^{\hat{L}}$, and $E = \{\{e_{tr,ij}\}_{i,j=1}^{N}\}_{t,r=1}^{\hat{L}}$. $E$ is denoted as the adjacent matrix of the FC graph, whose elements represent the correlations between sensors among all patches.
The graph $\mathcal{G}$ encompasses not only temporal correlations between timestamps and spatial correlations in each timestamp, but also additionally includes the correlations between DEDT, enabling us to model the comprehensive ST dependencies within MTS data.

\subsubsection{Decay Matrix}

The FC graph $\mathcal{G}$ is constructed based on sensor similarity across patches only, without accounting for temporal distances between sensors across these patches. However, it is intuitive that sensors at more distant timestamps should show weaker correlations compared to those at closer timestamps. Motivated by this, we devise a decay matrix that incorporates temporal distances between sensors, aiming to enhance the precision of the FC graph $\mathcal{G}$.

\begin{figure}[H]
    \centering\includegraphics[width = .75\linewidth]{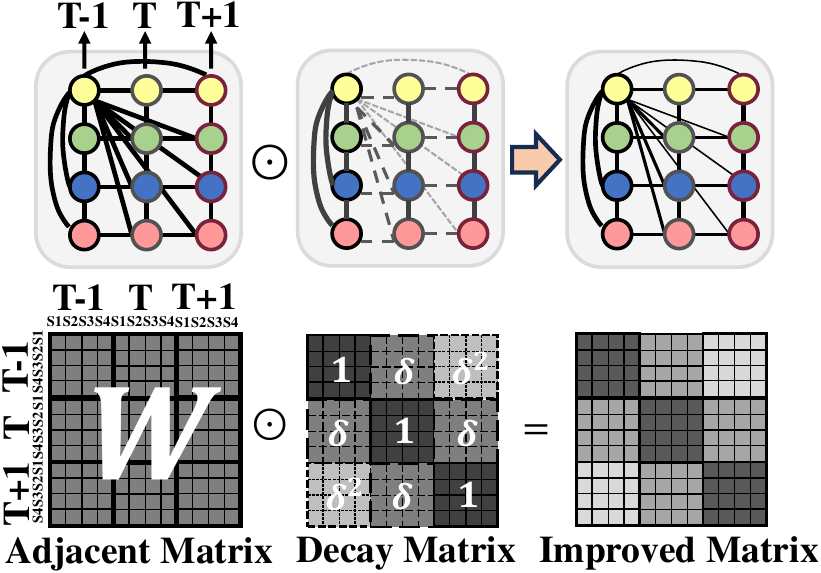}
    \caption{Decay matrix to improve the adjacent matrix.}
    \label{fig:decay}
\end{figure}

We provide Fig. \ref{fig:decay} for visual clarification. The left represents the adjacency matrix of a graph involving three patches, each containing four sensors. The dimension of this adjacent matrix is $E\in\mathbb{R}^{(3\times4)\times(3\times4)}$. In this matrix, each row presents a sensor's connections with other sensors across all patches. We take the first row as an example, which represents the connectivity of the first sensor $z_{T-1,1}$ of the $(T-1)$-th patch. The first four columns represent its connections with sensors within the same patch. As these sensors occur simultaneously, they should exhibit stronger correlations than those in other patches. The subsequent four columns represent the connections of $z_{T-1,1}$ with sensors from the $T$-th patch. As these sensors are in different patches, their correlations with $z_{T-1,1}$ should be decayed, measured by a decay rate $\delta$. The final four columns represent the connections of $z_{T-1,1}$ with sensors from the $(T+1)$-th patch. As the temporal gap expands, correlations naturally decline further, measured by $\delta^2$.
Drawing from these discussions, we formulate the decay matrix $C = \{\{c_{tr,ij}\}_{i,j=1}^{N}\}_{t,r=1}^{\hat{L}}$, where each element $c_{tr,ij} = \delta^{t-r}$. This matrix is employed to enhance the correlations between sensors across patches, yielding $e_{tr, ij} = e_{tr, ij}\cdot{c_{tr,ij}}$. This approach ensures that temporally close sensors exhibit stronger correlations than those temporally distant sensors.

\subsection{FC Graph Convolution}
Utilizing the constructed FC graph, the next step is to capture the ST dependencies within MTS data for representation learning. A straightforward approach would involve applying graph convolution across the entire graph. Nevertheless, this approach might fail to effectively capture the local ST dependencies within MTS data. This is similar to the rationale behind CNNs employing local convolution to capture local information from images. Furthermore, directly utilizing the entire graph could lead to extra computation costs. 
To solve these limitations, we propose a moving-pooling GNN, including a moving window to capture local ST dependencies and temporal pooling to extract high-level features.

\begin{figure}[htbp!]
    \centering\includegraphics[width = .55\linewidth]{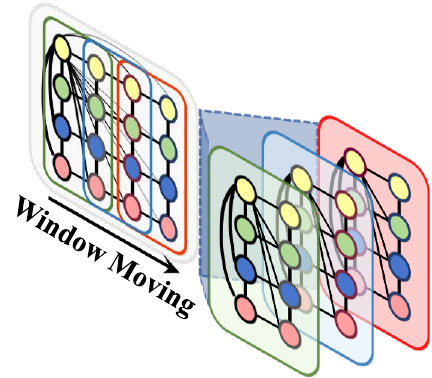}
        \caption{Three windows obtained by moving along patches.}
    \label{fig:pool_graph}
\end{figure}

We begin by utilizing a moving window with a specific size $M$ that traverses along patches. The window moves by $s$ slides in each movement. Fig. \ref{fig:pool_graph} provides a visual illustration, featuring an FC graph with four patches, each containing four sensors. In this example, a size-two window moves with stride one, leading to three windows obtained. Here, each window contains two patches, each containing four sensors.
Then, GNN is adopted within each window. 

Specifically, following previous works \cite{wang2023multivariate,deng2021graph}, we employ a Message Passing Neural Network (MPNN), a variant of GNN, to capture ST dependencies of the graph within each window. Specifically, MPNN involves propagation and updating stages. During the propagation stage, the information from neighboring nodes is propagated into the central node. Given a central node $z^l_{t,i}$ of the $w$-th window in the $l$-th layer, it has a set of neighboring nodes $\{\{z^l_{r,j}\}_{j=1}^N\}_{r=w-\frac{M}{2}}^{w+\frac{M}{2}}$ across $M$ patches in the same window. The central node has correlations with its neighbors as $\{\{e^l_{tr,ij}\}_{j=1}^{N}\}_{r=w-\frac{M}{2}}^{w+\frac{M}{2}}$. After the propagation stage, we obtain the propagated features $h^l_{t,i} = \sum_{r=w-\frac{M}{2}}^{w+\frac{M}{2}}\sum^N_{j=1}z^l_{r,j}{e^l_{tr,ij}}$. Then, the updating stage adopts a non-linear function to update the propagated sensor features, i.e., $z_{t,i}^{l+1} = f_g(h^l_{t,i}|W_g)$. 
Overall, MPNN propagates the information of sensors based on the correlations between all sensors across $M$ patches, enabling us to fully capture the comprehensive ST dependencies within the window to update sensor features. The updating stage introduces non-linear functions to update sensor features, further enhancing the ability to learn effective representations.

After updating sensor features by capturing ST dependencies, a temporal pooling operation is employed to extract high-level features for each window, drawing inspiration from the pooling operation in CNNs. Given the updated sensor features $\{z_{t,i}^{l+1}\}_{t=w-\frac{M}{2}}^{w+\frac{M}{2}}$ for the $i$-the sensor across $M$ patches, we perform temporal pooling using an average pooling strategy, yielding sensor features $z_{w,i}^{l+1} = \sum_{t=w-\frac{M}{2}}^{w+\frac{M}{2}}z_{t,i}^{l+1}/{M}$ for the $w$-th window. Subsequently, by stacking the sensors across all windows as depicted in Fig. \ref{fig:overall}, we create a high-level FC graph serving as input for the subsequent layer. Note that we only adopt one layer in this study, thus directly utilizing the obtained sensor features from each window for output purposes.

Inspired by the multi-branch concept introduced in previous research \cite{vaswani2017attention}, we also integrate multiple parallel layers of graph construction and convolution. This approach allows us to initialize the model with diverse weights, enabling training to capture ST dependencies from various comprehensive viewpoints and obtain the best possible solution. Stacking all sensor features from these multiple layers, we employ a straightforward output layer, i.e., MLP, to transform the stacked features into representations. These representations can be leveraged for downstream tasks.

\begin{table*}[htbp]
  \centering
  \small
    \begin{tabular}{l|cccccccc}
    \toprule
    \toprule
    \multicolumn{1}{c|}{\multirow{2}[2]{*}{Models}}
          & \multicolumn{2}{c}{FD001} & \multicolumn{2}{c}{FD002} & \multicolumn{2}{c}{FD003} & \multicolumn{2}{c}{FD004} \\
          & RMSE  & Score & RMSE  & Score & RMSE  & Score & RMSE  & Score  \\
    \midrule
    AConvLSTM & 13.10$\pm$0.37 & 286$\pm$45   & 13.11$\pm$0.21 & 737$\pm$65   & 12.13$\pm$0.53 & 276$\pm$75   & 14.64$\pm$0.31 & 1011$\pm$107   \\
    DAGN  & 16.11$\pm$0.21 & 595$\pm$131   & 16.43$\pm$0.05 & 1242$\pm$116  & 18.05$\pm$0.25 & 1216$\pm$177  & 19.04$\pm$0.10 & 2321$\pm$105   \\
    InFormer & 13.13$\pm$0.22 & 263$\pm$19   & 13.20$\pm$0.15 & 715$\pm$71   & 12.58$\pm$0.24 & 228$\pm$15   & 14.16$\pm$0.49 & 1023$\pm$201   \\
    AutoFormer & 23.04$\pm$0.28 & 1063$\pm$73  & 16.51$\pm$0.47 & 1248$\pm$112  & 25.40$\pm$0.26 & 2034$\pm$163  & 20.31$\pm$0.14 & 2291$\pm$122   \\
    GCN   & \underline{12.58$\pm$0.22} & 237$\pm$24   & 13.78$\pm$0.22 & 849$\pm$62   & \underline{11.92$\pm$0.15} & \underline{218$\pm$33}   & 14.44$\pm$0.32 & 967$\pm$66    \\
    HAGCN & 13.10$\pm$0.63 & 263$\pm$30   & 14.92$\pm$0.12 & 1086$\pm$87  & 13.46$\pm$0.30 & 327$\pm$52   & 14.66$\pm$0.25 & 880$\pm$150    \\
    HierCorrPool & 12.64$\pm$0.23 & \underline{227$\pm$21}   & 13.23$\pm$0.31 & \textbf{709$\pm$61} & 12.30$\pm$0.15 & 220$\pm$16   & \underline{13.86$\pm$0.32} & \underline{854$\pm$68}    \\
    MAGNN & 12.63$\pm$0.32 & 246$\pm$25   & \underline{13.09$\pm$0.13} & \underline{714$\pm$57}   & 12.15$\pm$0.16 & 253$\pm$32   & 14.30$\pm$0.26 & 978$\pm$137    \\
    \midrule
    Ours  & \textbf{11.62$\pm$0.19} & \textbf{203$\pm$16} & \textbf{13.04$\pm$0.13} & 738$\pm$49   & \textbf{11.52$\pm$0.19} & \textbf{198$\pm$12} & \textbf{13.62$\pm$0.25} & \textbf{816$\pm$63}  \\
    \bottomrule
    \bottomrule
    \end{tabular}%
    \caption{Comparisons with SOTAs in C-MAPSS}
      \label{tab:sotarul}%
\end{table*}%

\begin{table}[htbp]
\small
  \centering
    \begin{tabular}{l|cccc}
    \toprule
    \toprule
    \multicolumn{1}{c|}{\multirow{2}[2]{*}{Models}} & \multicolumn{2}{c}{{UCI-HAR}}  \\
          & Accu  & MF1   \\
    \midrule
    AConvLSTM  & 86.06$\pm$1.01 & 85.75$\pm$1.01  \\
    DAGN   & 89.02$\pm$0.49 & 88.94$\pm$0.48  \\
    InFormer  & 90.23$\pm$0.48 & 90.23$\pm$0.47  \\
    AutoFormer  & 56.70$\pm$0.81 & 54.41$\pm$1.74  \\
    GCN    & \underline{94.79$\pm$0.33} & \underline{94.82$\pm$0.33}  \\
    HAGCN  & 80.79$\pm$0.77 & 81.08$\pm$0.75  \\
    HierCorrPool  & 93.81$\pm$0.26 & 93.79$\pm$0.28  \\
    MAGNN   & 90.91$\pm$0.99 & 90.79$\pm$1.08  \\
    \midrule
    Ours   & \textbf{95.81$\pm$0.24} & \textbf{95.82$\pm$0.24}  \\
    \bottomrule
    \bottomrule
    \end{tabular}%
    \caption{Comparisons with SOTAs in UCI-HAR}
      \label{tab:sotahar}%
\end{table}%

\section{Experimental Results}
\paragraph{Datasets}
We examine our method on three different downstream tasks: Remaining Useful Life (RUL) prediction, Human Activity Recognition (HAR), and Sleep Stage Classification (SSC). Specifically, we utilize C-MAPSS \cite{saxena2008CMAPSS} for RUL prediction, UCI-HAR \cite{anguita2012human} for HAR, and ISRUC-S3 \cite{khalighi2016isruc} for SSC, following the previous work \cite{wang2023multivariate}. For C-MAPSS which includes four sub-datasets, we adopt the pre-defined train-test splits. The training dataset is further divided into 80\% and 20\% for training and validation. For HAR and ISRUC, we randomly split them into 60\%, 20\%, and 20\% for training, validating, and testing. The details of these datasets can be found in our appendix. 

\paragraph{Evaluation}
To evaluate the performance of RUL prediction, we adopt RMSE and the Score function, following previous works \cite{chen2020machine,wang2023sensor}. Lower values of these indicators refer to better model performance. For the evaluation of HAR and SSC, we adopt Accuracy (Accu.) and Macro-averaged F1-Score (MF1) in accordance with prior studies \cite{eldele2021time,meng2022mhccl}. Larger values of these indicators refer to better performance. Besides, to reduce the effect of random initialization, we conduct ten times for all experiments and take the average results for comparisons. 

\paragraph{Implementation Details}
All methods are conducted with NVIDIA GeForce RTX 3080Ti and implemented by PyTorch 1.9. We set the batch size as 100, choose ADAM as the optimizer with a learning rate of 1e-3, and train the model 40 epochs. More details can be found in our appendix.

\subsection{Comparisons with State-of-the-Art}
We compare our method with SOTA methods, encompassing conventional methods like AConvLSTM \cite{xiao2021dual}, DAGN \cite{8723466}, Transformer-based approaches such as InFormer \cite{zhou2021informer} and AutoFormer \cite{wu2021autoformer}, as well as GNN-based methods including GCN \cite{kipf2016semi}, HAGCN \cite{LI2021107878}, HierCorrPool \cite{wang2023multivariate}, and MAGNN \cite{chen2023multi}. All methods are re-implemented based on their original configurations, with the exception of GNN-based methods, where we replace their encoders with the same encoders used in our approach for fair comparison.

Table \ref{tab:sotarul}, \ref{tab:sotahar}, and \ref{tab:sotassc} present the comparison results, showing the remarkable effectiveness of FC-STGNN. As shown in the tables, our method exhibits large improvements across a majority of cases in comparison to both conventional temporal encoder-based and GNN-based methods. For instance, our method shows improvements of 7.6\% and 3.4\% in FD001 and FD003 of C-MAPSS, respectively, over the second-best results regarding RMSE. Similar improvements can be observed in UCI-HAR and ISRUC-S3, where our method outperforms second-best methods by 1.02\% and 1.56\% regarding accuracy, respectively. These advancements underline the necessity of fully capturing spatial-temporal dependencies within MTS data, thus enabling superior performance compared to SOTA methods.

\begin{table}[htbp]
\small
  \centering
    \begin{tabular}{l|cccc}
    \toprule
    \toprule
    \multicolumn{1}{c|}{\multirow{2}[2]{*}{Models}} &  \multicolumn{2}{c}{{ISRUC-S3}} \\
          & Accu  & MF1   \\
    \midrule
    AConvLSTM  & 72.93$\pm$0.62 & 69.52$\pm$1.00 \\
    DAGN   & 55.35$\pm$0.35 & 50.51$\pm$2.78 \\
    InFormer  & 72.15$\pm$2.41 & 68.67$\pm$3.42 \\
    AutoFormer  & 43.75$\pm$0.95 & 37.88$\pm$2.43 \\
    GCN    & 79.62$\pm$0.38 & 77.57$\pm$0.94 \\
    HAGCN  & 66.59$\pm$0.29 & 60.20$\pm$2.24 \\
    HierCorrPool  & \underline{79.31$\pm$0.60} & \underline{76.25$\pm$0.72} \\
    MAGNN   & 68.13$\pm$2.54 & 64.31$\pm$5.25 \\
    \midrule
    Ours   & \textbf{80.87$\pm$0.21} & \textbf{78.79$\pm$0.55} \\
    \bottomrule
    \bottomrule
    \end{tabular}%
    \caption{Comparisons with SOTAs in ISRUC-S3}
      \label{tab:sotassc}%
\end{table}%

\subsection{Ablation Study}

\begin{table*}[htbp]
  \centering
  \small
    \begin{tabular}{l|cccccccc}
    \toprule
    \toprule
    \multirow{2}[2]{*}{Variants} 
          & \multicolumn{2}{c}{FD001} & \multicolumn{2}{c}{FD002} & \multicolumn{2}{c}{FD003} & \multicolumn{2}{c}{FD004}  \\
          & RMSE  & Score & RMSE  & Score & RMSE  & Score & RMSE  & Score \\
    \midrule
    w/o FC GC$^2$ & 12.58$\pm$0.22 & 237$\pm$24   & 13.78$\pm$0.22 & 849$\pm$62   & 11.92$\pm$0.15 & 218$\pm$33   & 14.44$\pm$0.32 & 967$\pm$66    \\
    w/o M\&P & 12.17$\pm$0.16 & 217$\pm$23   & 13.29$\pm$0.13 & 769$\pm$69   & 11.75$\pm$0.19 & 219$\pm$31   & 14.03$\pm$0.25 & 837$\pm$64    \\
    w/o Pooling & 12.03$\pm$0.22 & 231$\pm$24   & 13.13$\pm$0.22 & 720$\pm$61   & 11.68$\pm$0.15 & 220$\pm$33   & 13.74$\pm$0.24 & 849$\pm$58    \\
    w/o decay & 12.15$\pm$0.17 & 233$\pm$21   & 13.20$\pm$0.21 & 750$\pm$58   & 11.74$\pm$0.17 & 205$\pm$25   & 13.86$\pm$0.26 & 853$\pm$71    \\
    \midrule
    Complete & 11.62$\pm$0.19 & 203$\pm$16 & 13.04$\pm$0.13 & 738$\pm$49   & 11.52$\pm$0.19 & 198$\pm$12 & 13.62$\pm$0.25 & 816$\pm$63  \\
    \bottomrule
    \bottomrule
    \end{tabular}%
    \caption{Ablation study in C-MAPSS}
        \label{tab:ablarul}%
\end{table*}%
\begin{table}[htbp]
  \centering
  \small
    \begin{tabular}{l|cccc}
    \toprule
    \toprule
    \multirow{2}[2]{*}{Var. w/o} &  \multicolumn{2}{c}{{UCI-HAR}} & \multicolumn{2}{c}{{ISRUC-S3}} \\
           & Accu  & MF1   & Accu  & MF1 \\
    \midrule
     FC GC$^2$   & 94.79$\pm$0.33 & 94.82$\pm$0.33 & 79.62$\pm$0.38 & 77.57$\pm$0.94 \\
     M\&P   & 95.06$\pm$0.26 & 95.10$\pm$0.26 & 79.85$\pm$0.27 & 77.60$\pm$0.74 \\
     Pooling    & 95.53$\pm$0.30 & 95.57$\pm$0.30 & 80.16$\pm$0.32 & 77.42$\pm$0.80 \\
     decay    & 95.20$\pm$0.27 & 95.28$\pm$0.28 & 80.13$\pm$0.32 & 78.33$\pm$0.72 \\
    \midrule
    Comp.  & {95.81$\pm$0.24} & 95.82$\pm$0.24 & 80.87$\pm$0.21 & 78.79$\pm$0.55 \\
    \bottomrule
    \bottomrule
    \end{tabular}%
    \caption{Ablation study in UCI-HAR and ISRUC-S3}
    \label{tab:ablaharssc}%
\end{table}%
We conducted an ablation study to assess the effectiveness of our proposed modules. In the first variant 'w/o FC GC$^2$', we excluded the usage of our FC Graph Construction and Graph Convolution approach. Instead, we followed the conventional methods \cite{ijcai2020-184,wang2023multivariate} to separately construct and convolve graphs for each patch. The second variant 'w/o M\&P' involved incorporating FC graph construction but omitted the moving window and temporal pooling, so local ST dependencies cannot be captured. Furthermore, we obtained the third variant 'w/o pooling' by introducing the moving window while excluding the temporal pooling operation that is designed for high-level features. Lastly, the 'w/o decay' variant refrained from using the designed decay matrix to enhance the constructed FC graph. These variants are compared with the complete version.

Table \ref{tab:ablarul} and \ref{tab:ablaharssc} present the ablation study across three datasets. We take the RMSE results on FD001 of C-MAPSS as examples. Comparing against the 'w/o FC GC$^2$' variant, we observe that our complete method achieves a 7.6\% improvement, highlighting the necessity of the FC graph for effective feature learning through comprehensive modelling of ST dependencies within MTS data. With the introduction of FC graph construction, there is a noticeable performance boost of the 'w/o M\&P' variant, and the gap with the complete version narrows, i.e., the gap is reduced to 4.5\%. This outcome suggests that the FC graph contributes to representation learning, even without accounting for local ST dependencies within MTS data.
Furthermore, by incorporating the moving window approach, we witness further performance improvements of the 'w/o pooling' variant due to its effectiveness in capturing local ST dependencies, narrowing the gap to 3.4\%. With the inclusion of the temporal pooling operation, high-level sensor features are obtained, which helps to eliminate redundant features and thus further enhance the performance. Finally, when the decay matrix is excluded, there is a 4.3\% decrease in performance, emphasising the necessity of employing the decay matrix to refine the constructed FC graph.

The above observations hold true across other sub-datasets of C-MAPSS, UCI-HAR, and ISRUC-S3 as well. These results underline the importance of modelling the comprehensive ST dependencies within MTS data, which in turn allows for the learning of more effective representations. This comprehensive modelling leads to superior overall performance in various downstream tasks.

\subsection{Sensitivity Analysis}
In this section, we conduct sensitivity analysis for No. of parallel layers, patch size, moving window size, and decay rate. Typical results are reported, and additional results can be found in our appendix.
\subsubsection{No. of Parallel Layers}
\begin{figure}[htbp!]
    \centering\includegraphics[width = .9\linewidth]{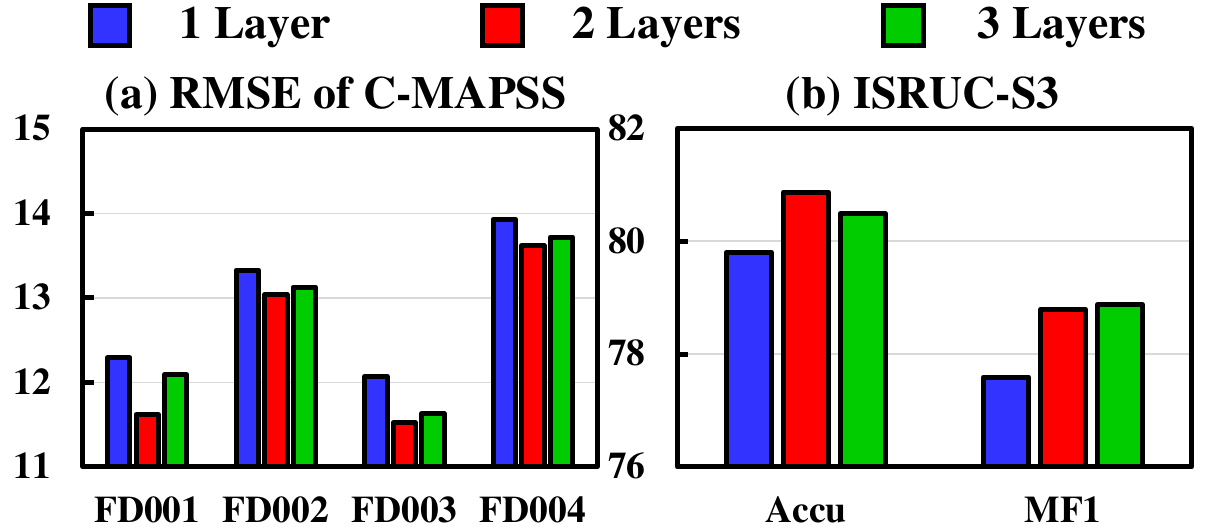}
    \caption{Sensitivity analysis for No. of parallel layers.}
    \label{fig:layer}
\end{figure}
In our approach, we employ multiple parallel layers of FC graph construction and graph convolution, allowing us to capture the spatial-temporal dependencies within MTS data from diverse perspectives. To assess the impact of varying the number of layers, we obtain the results in Fig. \ref{fig:layer}. It can be observed that incorporating additional parallel layers leads to enhanced performance, affirming the efficacy of employing multiple layers to model ST dependencies. For instance, in all cases, the model with 2 layers outperforms the single-layer counterpart. Additionally, in specific cases of ISRUC-S3, introducing 3 layers contributes to better performance compared to using fewer layers. However, the performance gains start diminishing or even reversing when many layers are introduced due to overfitting. Thus, too many layers are unnecessary.

\subsubsection{Patch Size Analysis}
\begin{figure}[htbp!]
    \centering\includegraphics[width = .9\linewidth]{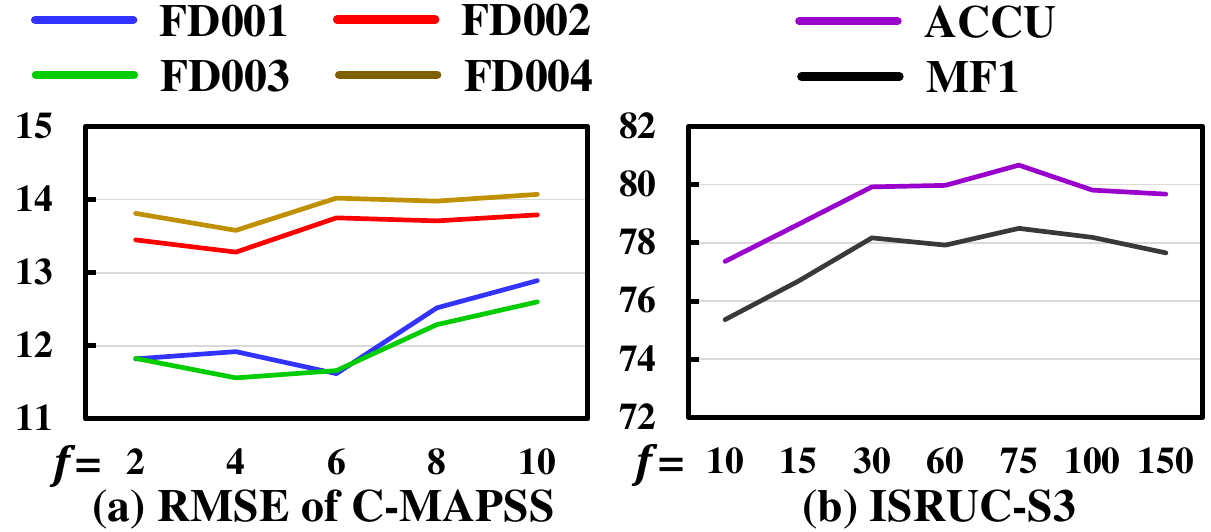}
    \caption{Sensitivity analysis for patch sizes.}
    \label{fig:patchsizerul}
\end{figure}
We segment each MTS sample as multiple patches for FC graph construction, which makes the patch size a parameter $f$ influencing the constructed FC graph. To evaluate its impact, we conducted the patch size analysis. Notably, since C-MAPSS samples have relatively short time lengths, e.g., 30 timestamps for FD001, we opted for smaller patch sizes within [2, 4, 6, 8, 10] for sample segmentation. While for those in ISRUC-S3 which have larger time lengths, i.e., 300, we explored patch sizes within [10, 15, 30, 60, 75, 100, 150].

Fig. \ref{fig:patchsizerul} presents the results. For C-MAPSS where sample sizes are small, we find that relatively smaller patch sizes would be good to obtain better performance. For instance, considering the RMSE of FD001, the optimal performance is achieved when the patch size is set to 6. Similar trends can be found across various sub-datasets, where the best performance can be generally found when the patch sizes are set to 4 or 6. Conversely, for datasets characterized by larger time lengths, employing relatively larger patch sizes leads to improved performance. For example, ISRUC-S3 samples exhibit enhanced performance with patch sizes around 75. These observations emphasize the nuanced relationship between patch size and performance, which is influenced by the characteristics of a specific dataset.

\begin{table*}[htbp]
  \centering
  \small
    \begin{tabular}{c|cccc|cccc}
    \toprule
    \toprule
    \multirow{2}[2]{*}{Indicators} & \multicolumn{4}{c|}{FD001 of C-MAPSS} & \multicolumn{4}{c}{ISRUC-S3} \\
    \multicolumn{1}{c|}{} & \multicolumn{1}{c}{FLOPs} & \# Weights & Training /s & Inference /ms & FLOPs & \# Weights & Training /s & Inference /ms \\
    \midrule
    GCN   & 1,793,288 & 38,625 & 73    & 2.34  & 17,194,912 & 111,662 & 242   & 2.78 \\
    HAGCN & 1,843,336 & 22,436 & 78    & 2.09  & 17,150,632 & 197,828 & 213   & 4.04 \\
    HierCorr & 2,717,034 & 1,071,906 & 89    & 2.34  & 24,998,334 & 7,929,590 & \textbf{194}   & 2.65 \\
    MAGNN & 2,181,150 & 30,464 & 453   & 5.61  & 19,280,332 & 155,923 & 215   & 10.31 \\
    \midrule
    Ours  & \textbf{856,072} & \textbf{20,225} & \textbf{68}    & \textbf{2.03}  & \textbf{15,422,112} & \textbf{51,822} & 210   & \textbf{2.23} \\
    \bottomrule
    \bottomrule
    \end{tabular}%
    \caption{Comparisons of model complexity}
      \label{tab:complex}%
\end{table*}%
\subsubsection{Moving Window Size Analysis}
\begin{figure}[htbp!]
    \centering\includegraphics[width = .9\linewidth]{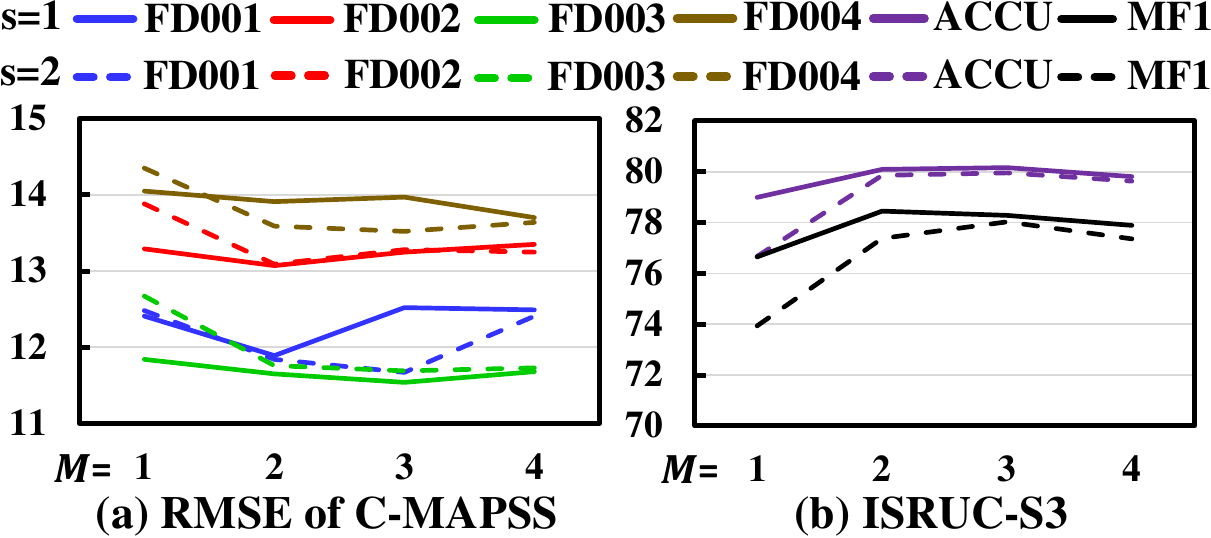}
    \caption{Sensitivity analysis for moving window sizes.}
    \label{fig:kernelsizerul}
\end{figure}
We utilize moving windows with a designated size $M$, which traverse along the patches with stride $s$, to capture the local ST dependencies within MTS data. To evaluate their effects, we consider window sizes $M$ of [1, 2, 3, 4], and stride sizes $s$ of [1, 2].

Fig. \ref{fig:kernelsizerul} shows the analysis results. We consider the RMSE on C-MAPSS as examples. We find that a larger $M$ can help to obtain better performance. For instance, the variant with $M=2$ outperforms those with $M=1$ which represents the variant without considering the correlations between DEDT. The improvements highlight the importance of considering these correlations through our FC graph. Additionally, further increasing $M$ does not consistently yield additional benefits. In fact, performance may decrease when $M$ becomes too large, e.g., $M=4$ for FD001. This is because larger $M$ includes more patches within each window for graph convolution, potentially causing local ST dependencies to be poorly captured. Meanwhile, similar trends can be found when $s=2$. Notably, the performance of $s=2$ is generally poorer when $M$ is smaller, as small $M$ and large $s$ will lose information when moving the windows. Overall, these findings suggest that a window size of $M=2$ and stride size $s=1$ are optimal for achieving the best performance.

\subsubsection{Decay Rate Analysis}
\begin{figure}[htbp!]
    \centering\includegraphics[width = .9\linewidth]{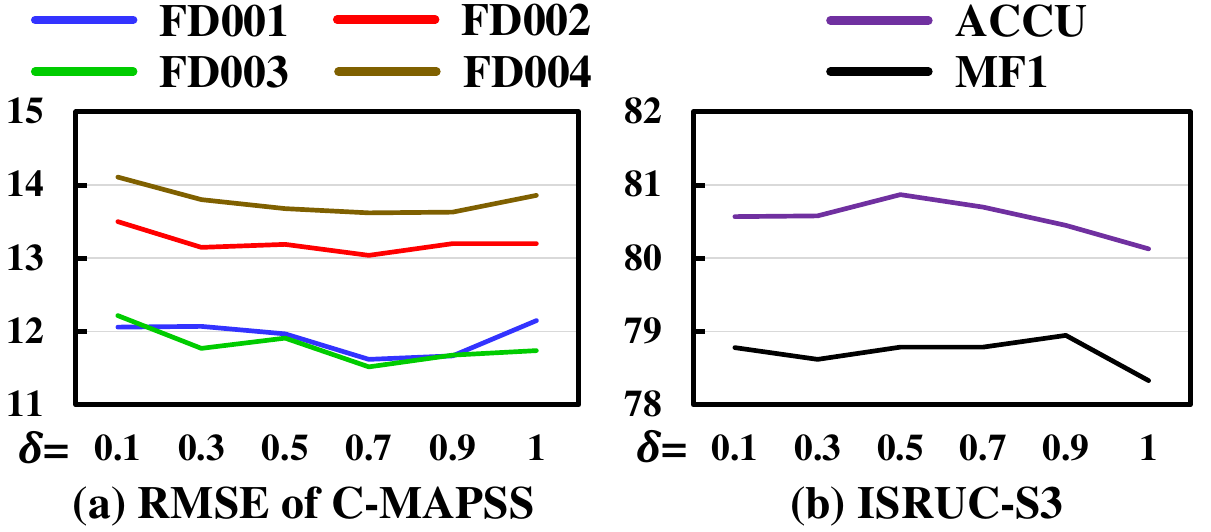}
    \caption{Sensitivity analysis for decay rates.}
    \label{fig:decayrul}
\end{figure}
We employ the decay matrix to enhance our FC graph for accurately representing the correlations between DEDT. The choice of decay rate $\delta$ is crucial and thus necessitates evaluation. We consider $\delta$ values within [0.1, 0.3, 0.5, 0.7, 0.9, 1], where $\delta=1$ represents the variant without using the decay matrix. From the results in Fig. \ref{fig:decayrul}, we find that the variants with relatively larger $\delta$ yield better performance, such as $\delta=0.7$ and $\delta=0.9$. When $\delta$ is exceedingly small, e.g., 0.1, the performance experiences a significant drop, as the correlations between DEDT are overly distorted. Thus, setting $\delta$ to 0.7 or 0.9 proves effective in achieving good performance for our model.

\subsection{Model Complexity}

Model complexity is a critical factor in determining a model's applicability to real systems. Excessive complexity, even if it yields good performance, may render a model impractical. In this section, we conduct a comprehensive comparison of our method with four highly competitive approaches. The evaluation includes Floating-point Operations Per Second (FLOPs) and model weights, representing time complexity and the number of trainable weights, respectively. Additionally, we compare the training and inference times to assess the real-time requirements during these processes. The training time is measured by training a model until convergence. For the inference time evaluation, we simulate the process in real systems by recording the time required to predict one sample at a time. To ensure a fair comparison, all methods are executed on the same computation platform. The results of the comparisons are presented in Table \ref{tab:complex}. We conducted the evaluations in two scenarios, i.e., RUL prediction (C-MAPSS) and SSC (ISRUC-S3). The findings suggest that our method exhibits reasonable model complexity compared to SOTA approaches. Notably, our method requires the fewest FLOPs and trainable weights, indicating its suitability for deployment in real systems. Furthermore, in terms of inference, our method demonstrates the lowest inference time, emphasizing its practicality. Although our method requires slightly more training time compared to HierCorr (210 vs. 194), the difference is marginal.

\section{Conclusion}
To model the comprehensive Spatial-Temporal (ST) dependencies within MTS data, we design a novel method named as Fully-Connected Spatial-Temporal Graph Neural Network (FC-STGNN). The method includes two essential modules, FC graph construction and FC graph convolution. For graph construction, we design an FC graph to connect sensors among all timestamps by additionally considering the correlations between DEDT, enabling comprehensive ST dependencies modelling within MTS data. Next, FC graph convolution is designed, with a moving-pooling GNN by leveraging a moving window and temporal pooling to capture the local ST dependencies and then learn high-level features. Our method is evaluated through extensive experiments, emphasizing its capacity to effectively model the comprehensive ST dependencies within MTS data.
\section*{Acknowledgements}
We thank anonymous reviewers for their constructive comments on this work. This research is supported by the Agency for Science, Technology and Research (A*STAR) under its AME Programmatic Funds (Grant No. A20H6b0151) and Career Development Award (Grant No. C210112046), and the National Research Foundation, Singapore under its AI Singapore Programme (AISG2-RP-2021-027).
\bibliography{aaai24}

\end{document}